\documentclass[11pt,a4paper]{article}
\usepackage[final]{acl} 
\usepackage{times}
\usepackage{latexsym}

\usepackage[colorinlistoftodos]{todonotes}
\usepackage{times}
\usepackage{latexsym}
\usepackage{graphicx}
\usepackage{booktabs}
\usepackage{float}
\usepackage{pgfplots}
\usepackage{pgf-pie}
\usepackage{hyperref}
\usepackage{listings}
\usepackage{listingsutf8}
\usepackage{comment}
\usepackage{xcolor}
\usepackage{float}
\usepackage{url} 
\pgfplotsset{compat=1.18}
\definecolor{codegreen}{rgb}{0,0.6,0}
\definecolor{codegray}{rgb}{0.5,0.5,0.5}
\definecolor{codepurple}{rgb}{0.58,0,0.82}
\definecolor{backcolour}{rgb}{0.95,0.95,0.92}

\usepackage[T1]{fontenc}
\usepackage[utf8]{inputenc}

\usepackage[romanian, english]{babel}
\lstdefinelanguage{json}{
    basicstyle=\ttfamily\footnotesize,
    numbers=none,
    stepnumber=1,
    numbersep=8pt,
    showstringspaces=false,
    breaklines=true,
    frame=single,
    backgroundcolor=\color{white},
    literate=
     *{0}{{{\color{blue}0}}}{1}
      {1}{{{\color{blue}1}}}{1}
      {2}{{{\color{blue}2}}}{1}
      {3}{{{\color{blue}3}}}{1}
      {4}{{{\color{blue}4}}}{1}
      {5}{{{\color{blue}5}}}{1}
      {6}{{{\color{blue}6}}}{1}
      {7}{{{\color{blue}7}}}{1}
      {8}{{{\color{blue}8}}}{1}
      {9}{{{\color{blue}9}}}{1}
      {:}{{{\color{black}:}}}{1}
      {,}{{{\color{black},}}}{1}
      {\{}{{{\color{red}\{}}}{1}
      {\}}{{{\color{red}\}}}}{1}
      {[}{{{\color{red}[}}}{1}
      {]}{{{\color{red}]}}}{1},
}

\lstdefinestyle{mystyle}{
    backgroundcolor=\color{backcolour},
    commentstyle=\color{codegreen},
    keywordstyle=\color{magenta},
    numberstyle=\tiny\color{codegray},
    stringstyle=\color{codepurple},
    basicstyle=\ttfamily\scriptsize, 
    breakatwhitespace=false,
    breaklines=true, 
    captionpos=b,
    keepspaces=true,
    numbers=none, 
    numbersep=5pt,
    showspaces=false,
    showstringspaces=false,
    showtabs=false,
    tabsize=2
}

\title{
    GRILE: A Benchmark for Grammar Reasoning and Explanation in Romanian LLMs
}
\author{ Adrian-Marius Dumitran\thanks{Equal contribution.} 
Alexandra-Mihaela Danila\footnotemark[1] Angela-Liliana Dumitran\\
University of Bucharest Faculty of Mathematics and Computer Science}

\begin{document}
\maketitle

\begin{abstract}
LLMs (Large language models) have revolutionized NLP (Natural Language Processing), yet their pedagogical value for low‑resource languages remains unclear. We present GRILE (Grammar Romanian Inference and Language Explanations) , the first open benchmark of 1,151 multiple‑choice questions harvested from Romanian high‑stakes exams (National Evaluation, Baccalaureate, university admissions). GRILE enables us to probe two complementary abilities of seven state‑of‑the‑art multilingual and Romanian‑specific LLMs: (i) selecting the correct answer, and (ii) producing linguistically accurate explanations. While Gemini 2.5 Pro reaches 83\% accuracy, most open‑weight models stay below 65\%, and 48\% of their explanations contain factual or pedagogical flaws according to expert review. A detailed error analysis pinpoints systematic weaknesses in morphology and in applying the latest DOOM3 orthographic norms. All data, code and a public web demo are released to catalyze future research. Our findings expose open challenges for trustworthy educational NLP in low‑resource settings and establish GRILE as a new test‑bed for controllable explanation generation and evaluation.
\end{abstract}

\section{Introduction}

Recent advancements in NLP, particularly through LLMs, have unlocked new opportunities for intelligent computer-assisted language learning. Although extensive benchmarks exist for English and other high-resource languages \citep{hendrycks2021measuringmassivemultitasklanguage,openai2024gpt4technicalreport}, their efficacy in low-resource contexts remains under-explored—especially for tasks that demand fine-grained grammatical reasoning. Romanian, spoken by roughly 24 million people worldwide, illustrates this gap: it is comparatively underserved by both annotated corpora and NLP research. \citep{Pais2023} mention that the number of identified NLP tools for Romanian represent only 15\% of the tools available for English.

This paper bridges that gap through an interdisciplinary collaboration between linguists and computer scientists. We introduce \textbf{GRILE}, a new benchmark of 1,151 single-answer multiple-choice questions (MCQs) extracted from high-stakes Romanian examinations, including the National Evaluation, the Baccalaureate, and several university entrance tests (e.g., Law School, Police Academy).

Our investigation pursues two principal goals. First, we quantitatively assess how accurately state-of-the-art multilingual and Romanian-specific LLMs answer the questions in our dataset. Second, we examine their pedagogical potential by requiring each model to justify its choice with a concise grammatical explanation. The evaluation covers commercial systems such as \textbf{Gemini 2.5 Pro Experimental} and open-weight models such as \textbf{DeepSeek V3-0324}.

Initial results reveal substantial variation, with answer accuracy spanning from \textbf{38\%} to \textbf{85\%}, and chain-of-thought prompting generally yielding higher scores. Beyond answer prediction, we instruct Gemini 2.5 Pro Experimental to categorise every question as \emph{Lexical}, \emph{Morphological}, \emph{Syntactic}, or \emph{Phonetic}. Manual verification confirms more than 99\% agreement, offering an additional lens for analysing error patterns. A qualitative evaluation of the generated explanations—conducted by a specialist in Romanian grammar—highlights persistent issues in terminological precision and normative compliance, underscoring the challenges of deploying LLMs as trustworthy tutors for low-resource languages.

Our main contributions are:

\begin{itemize}
    \item \textbf{GRILE}, the first public Romanian grammar benchmark paired with expert-validated explanations;
    \item a comprehensive evaluation of multilingual and Romanian-specific LLMs on both \emph{answering} and \emph{explaining} high-stakes grammar questions;
    \item a linguist-driven qualitative study that uncovers systematic weaknesses and normative misalignments;
    \item an open-access dataset and interactive web application, released under a permissive licence on Zenodo\footnote{\url{https://zenodo.org/records/15504323}}, to catalyse future research and educational use.
\end{itemize}

These resources and findings advance the study of educational NLP in low-resource settings and lay the groundwork for more reliable, Romanian-focused grammatical tutoring and assessment tools.

\section{Related Work}

\paragraph{Benchmarking LLMs on Multiple-Choice Grammar Tasks.}
Recent studies have shown that large language models (LLMs) can achieve impressive results on standardized multiple-choice tests across various domains \citep{hendrycks2021measuringmassivemultitasklanguage, openai2024gpt4technicalreport}. However, fine-grained grammatical assessments remain challenging for LLMs. For example, multilingual benchmarks focusing on grammar, such as the TELEIA \citep{spanish_benchmark_teleia} dataset for Spanish, demonstrate that LLMs struggle to reach native-like accuracy in nuanced grammar tasks \citep{mayor-rocher2024teleiaeval}. This motivates specialized evaluations of grammatical knowledge, especially for lower-resource languages like Romanian.

\paragraph{Multilingual vs. Language-Specific LLMs.}
Most large language models are predominantly trained on English data, which results in significantly better performance for English compared to other languages \citep{hu2020xtreme}. Consequently, this has encouraged the development of monolingual or language-specific NLP models that often outperform multilingual counterparts on local linguistic tasks. For instance, \citet{masala-etal-2020-robert} introduced RoBERT, a Romanian-specific BERT-based model, demonstrating superior results over multilingual BERT across various Romanian NLP tasks such as sentiment analysis, dialect identification, and diacritic restoration.

Similar outcomes have been consistently reported for other languages, including French, Polish, and Japanese, where monolingual transformer models regularly surpass general multilingual baselines in language-specific evaluations \citep{martin-etal-2020-camembert, kuratov-arkhipov-2019-adapting, conneau-etal-2020-unsupervised}. More recently, the NLP community has seen efforts toward developing large-scale LLMs tailored specifically for individual languages, such as Finnish GPT-style models \citep{luukkonen-etal-2023-fingpt} and Chinese LLaMA2-based models \citep{cui2023chinese}, achieving state-of-the-art performance on localized benchmarks.

In the Romanian context, \citet{masala2024openllmrotechnicalreport} released RoLLaMA2, the first open-source Romanian-specific large language model based on LLaMA-2, trained on curated Romanian corpora and fine-tuned through instruction tuning. RoLLaMA2 achieved leading performance across multiple Romanian NLP benchmarks, underscoring the effectiveness of language-specific training. Our study contributes to this growing body of literature by directly comparing multilingual models against Romanian-specific models like RoLLaMA2, focusing particularly on grammatical tasks to evaluate the effectiveness and viability of smaller, targeted language models against more extensive multilingual counterparts.
\paragraph{LLM Explanation Generation and Educational Feedback.}
Generating explanations for educational purposes extends beyond correctness, encompassing the quality and pedagogical soundness of explanations. Prior works in educational NLP formalized feedback comment generation tasks, highlighting the need for meaningful automated feedback \citep{nagata2019toward}. Recently, generative LLMs like GPT-4 demonstrated potential in automatically generating high-quality explanations in science education, closely matching human-generated feedback \citep{lopez-cortez-etal-2024-gmeg}. Human-in-the-loop validation remains crucial to ensure explanation accuracy and educational value.

Our work builds upon these directions, benchmarking Romanian-specific and general LLMs against a Romanian grammar multiple-choice dataset, examining common errors, and assessing explanation quality through expert validation. This contributes toward understanding LLM capabilities and limitations within educational contexts for low-resource languages.

\section{Dataset and Methodology}

In this section we explain how we collect the data and the models we decided to benchmark in our test.

\subsection{Data Collection and Structure}

\begin{figure}
    \centering
  \includegraphics[width=1.0\linewidth]{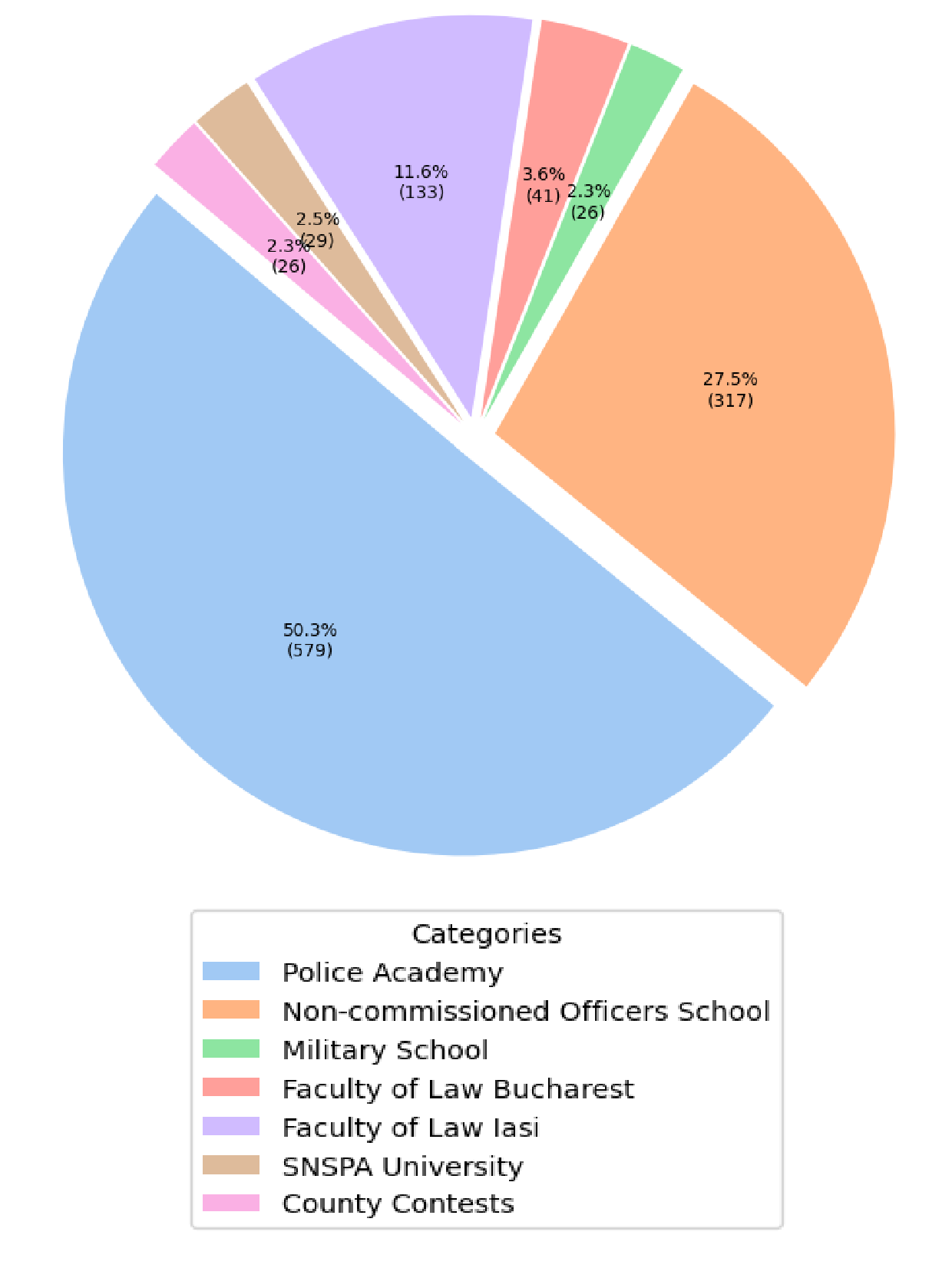}
    \caption{Dataset Sources}
    \label{fig:dataset_sources}
\end{figure}

Figure~\ref{fig:dataset_sources} charts the distribution of our 1\,151 single-answer multiple-choice questions (MCQs)  were harvested from publicly available high-stakes Romanian examinations—the \emph{National Evaluation}, \emph{Baccalaureate}, and university entrance tests for Law School and the Police Academy (2010–2024). Source PDFs and scans were converted with Tesseract OCR, then parsed by rule-based scripts and manually spot-checked. Each record is stored as JSON with fields \texttt{question}, \texttt{options}, \texttt{answer}, \texttt{source}, and \texttt{year}; underlined focus words are preserved via surrounding underscores. Full OCR pipeline details and an example entry appear in Appendix A.\footnote{\url{https://zenodo.org/records/15504323}}

\subsection{LLM Selection}
\label{sec:llm_selection}
Our model selection aimed to evaluate both a state-of-the-art model known for strong reasoning capabilities and several other prominent models readily accessible for research purposes. Specifically, we included \textbf{Gemini 2.5 Pro Experimental} \citep{geminiteam2024geminifamilyhighlycapable} as it represented the leading edge of publicly available models at the time of our experiments. The other benchmarked models, including variants of \textbf{DeepSeek}, \textbf{Mistral}, \textbf{Llama 3.3}, and \textbf{Qwen}, were chosen primarily for their strong performance in various benchmarks and their convenient accessibility via the Together AI API\footnote{\url{https://www.together.ai/}}, which facilitated systematic evaluation.

\begin{itemize}
    \item \textbf{DeepSeek V3-0324} \citep{liu2025deepseek}
    \item \textbf{Mistral Small 24B} 
    \item \textbf{Llama 3.3 70B Instruct} \citep{grattafiori2024llama}
    \item \textbf{Qwen 2.5 Coder 32B} \citep{yang2024qwen2}
    \item \textbf{Gemini 2.5 Pro Experimental},
    \textbf{Gemini 2.0 Flash} \citep{geminiteam2024geminifamilyhighlycapable} 
\end{itemize}

Furthermore, our few-shot prompting experiments (Section~\ref{sec:few_shot_results}) involved evaluating \textbf{multiple fine-tuned Romanian variants} (e.g., RoLlama, RoMistral, RoGemma based on OpenLLM-Ro \citep{masala2024openllmrotechnicalreport}) alongside their respective base models to analyze the interaction between fine-tuning and in-context learning. The full list of models and variants tested in the few-shot setting is provided in Appendix B.

\subsection{Prompting Strategies}
\label{sec:prompting}
We evaluated LLM performance using two primary prompting approaches applied to each MCQ in the dataset:

\subsubsection{Direct Questioning (Zero-Shot)}
This baseline approach involved providing the model directly with the question text and multiple-choice options, instructing it to select the correct answer. The prompt template was:
\begin{lstlisting}[frame=lines, basicstyle=\small\ttfamily, breaklines=true]
Question: [Question text here]
Choices:
(A) [Choice A text]
(B) [Choice B text]
(C) [Choice C text]
(D) [Choice D text, if applicable]
Select the correct answer. Ensure that the final answer is only the letter of the correct option, without any additional text or symbols.
\end{lstlisting}
The final instruction was added to minimize parsing errors during evaluation.

\subsubsection{Chain-of-Thought (CoT) Prompting}
To encourage step-by-step reasoning and generate explanations, we employed a CoT approach. The core CoT instruction added was:
\begin{quote}
\small\ttfamily
Let's think step by step. Provide your reasoning in several steps, and then output the final answer on a new line starting with "Final Answer:". Ensure that the final answer is only the letter of the correct option, without any additional text or symbols.
\end{quote}
This strategy aimed to improve answer accuracy by forcing a reasoning process and simultaneously generating explanatory text for qualitative analysis.

\subsubsection{Few-Shot Prompting}
To assess the impact of in-context examples, we also evaluated models using few-shot prompting. This involved prepending the Direct Questioning or CoT prompt (as applicable) with 1, 3, or 5 randomly selected question-answer pairs from the dataset (excluding the question being tested) formatted similarly to the main task.

\subsection{Explanation Generation and Validation Setup}

To facilitate efficient \textbf{linguistic expert validation} of both the answers and reasoning of the best-performing model identified in preliminary tests (Gemini 2.5 Pro Experimental), we used a simple, direct prompt asking it to provide its chosen answer, the correct answer (based on its knowledge), its reasoning, and a category classification. This generated a structured output specifically formatted for review, exemplified below:
\begin{lstlisting}[style=mystyle, % Apply the custom style
    caption={Example structured output from Gemini 2.5 Pro for validation.},
    label={lst:gemini_output_example},
    captionpos=b]
 Question 10: Din campul lexical al cuvantului "electorat" fac parte:
 Options:
   a) electron, neutron, proton
   b) electric, energetic, curent
   c) senator, primar, magistrat
   d) alegeri, vot, candidat
  Gemini's Answer: d
  Correct Answer: d
 Explanation: Campul lexical include cuvinte inrudite ca sens. "Electorat" se refera la totalitatea alegatorilor, fiind direct legat de "alegeri", "vot", "candidat".
 Category: Lexical
\end{lstlisting}

This structured format, including the category classification (Lexical, Morphological, Syntactic, Phonetic) assigned by Gemini 2.5 Pro Experimental, was used by the linguist expert to validate the model's answer, reasoning quality, and classification accuracy (which exceeded 99\% for the categorization task itself).

\subsection{Evaluation Metrics}
Performance was primarily measured by \textbf{answer accuracy}. Additionally, a \textbf{qualitative linguistic analysis} was conducted on the explanations generated via CoT (facilitated by the structured output from Gemini 2.5 Pro for validation) to assess their correctness, precision, and pedagogical suitability.

\section{Quantitative Results}
\label{sec:results}

This section presents quantitative findings from our benchmarking evaluations. We analyze the impact of different prompting strategies (Chain-of-Thought vs. Direct, Few-Shot vs. Zero-Shot) and compare the performance of Romanian-specific fine-tuned models against their multilingual base versions.

\subsection{Impact of Chain-of-Thought Prompting}
\label{sec:cot_results}

We first compared model accuracy using direct zero-shot questioning versus Chain-of-Thought (CoT) prompting, which instructs the model to provide reasoning steps. Table~\ref{tab:cot_comparison} summarizes these results.

\begin{table}[ht!]
\centering
\begin{tabular}{||p{3.3cm} p{1.6cm} p{1.5cm}||} 
 \hline
 Model & Accuracy without CoT & Accuracy with CoT \\ [0.5ex] 
 \hline\hline
 DeepSeek V3.0324 & \textbf{53.26\%} & 64.62\% \\ 
 Mistral small 24B & 39.10\% & 38.31\% \\ 
 Llama 3.3 70B & 49.70\% & 55.60\% \\ 
 Qwen 2.5 Coder 32B & 48.22\% & 51.61\% \\ 
\midrule 
 Gemini 2.5 Pro & ---     & \textbf{82.8\%} \\ 
 \hline
\end{tabular}
\caption{Comparison of model accuracy (\%) with and without Chain-of-Thought (CoT) prompting. Top score in each column is bolded. Gemini 2.5 Pro was evaluated only with CoT.}
\label{tab:cot_comparison}
\end{table}

 \textbf{CoT prompting} generally led to notable accuracy improvements, particularly for \texttt{DeepSeek V3} (+11.36 pp) and \texttt{Llama 3.3 70B} (+5.9 pp), confirming its benefit for reasoning on this task. An exception was \texttt{Mistral Small 24B}, which saw a slight decrease, possibly due to difficulties following complex instructions. \texttt{Gemini 2.5 Pro}, tested only with CoT, achieved the highest score (82.8\%), significantly outperforming others. However, accuracies for most models (38-65\% range with CoT) remain substantially below proficient human levels (>80\%) on comparable exams\footnote{Example results: \url{https://cdn.edupedu.ro/wp-content/uploads/2023/01/lista_anonimizata_rezultate_finale_LC.pdf}}, highlighting a remaining gap in nuanced Romanian grammar understanding for most LLMs.

\subsection{Impact of Few-Shot Prompting}
\label{sec:few_shot_results}

We also investigated whether providing in-context examples (1, 3, or 5 shots) could improve performance compared to zero-shot prompting. The results, detailed fully in Appendix B
, show \textbf{mixed and generally modest benefits}. While accuracy often increased slightly, gains were typically small (2-5 pp) and inconsistent across models and shot counts. Table~\ref{tab:few_shot_example} illustrates this variability for the Llama 3.1 8B family.

\begin{table}[ht]
 \centering
 \resizebox{\columnwidth}{!}{%
 \begin{tabular}{lcccc}
  \toprule
  \textbf{Llama 3.1 8B Variant} & \textbf{FS=0} & \textbf{FS=1} & \textbf{FS=3} & \textbf{FS=5} \\
  \midrule
  Base Instruct        & 35.62 & 38.84 & \textbf{40.75} & 39.70 \\
  Ro-Instruct (25-04)  & 36.06 & 37.71 & \textbf{40.49} & 39.36 \\
  Ro-Instruct-DPO (25-04) & 38.49 & 37.53 & \textbf{39.53} & 39.27 \\
  \bottomrule
 \end{tabular}%
 }
 \caption{Illustrative accuracy (\%) for Llama 3.1 8B variants with Few-Shot (FS) prompting. Best few-shot score per model in bold. Full results in Appendix B 
 .}
 \label{tab:few_shot_example}
\end{table}

\textbf{Few-shot prompting} did not consistently favor base models over fine-tuned ones, nor did a clear optimal number of shots emerge. Crucially, even with examples, peak accuracies for the models tested in this setting (detailed in Appendix B) 
remained significantly below top performers like Gemini 2.5 Pro and human levels. This suggests limited in-context examples are insufficient to overcome the core challenges of this task for these models.

\subsection{Performance of Romanian Fine-tuned Models vs. Base Models}
\label{sec:rollm_vs_base}

A key goal was to assess the impact of Romanian-specific fine-tuning. We compared the performance of various RoLLM variants (developed by \citet{masala2024openllmrotechnicalreport}) against their respective base multilingual models using data from our few-shot experiments (details in Appendix B).

The results indicate that Romanian-specific fine-tuning \textbf{does not consistently yield significant improvements} on this specific grammatical MCQ task, especially compared to strong base models.
\begin{itemize}
    \item For the \textbf{Llama} family (Llama-2, Llama-3, Llama-3.1), the RoLLM variants generally performed similarly to or sometimes slightly worse than their corresponding base Instruct models in zero-shot settings. Few-shot prompting occasionally led to RoLLM variants slightly outperforming their base, but the difference was often marginal. DPO fine-tuning on RoLLM variants did not produce consistent gains over standard RoLLM instruction tuning for this task.
    \item For \textbf{Mistral-7B}, the RoMistral variants showed slightly better average performance than the base Mistral-7B Instruct v0.3, but the difference was small (within 1 pp on average across FS settings).
    \item For \textbf{Gemma} models, the results were particularly striking. The base \texttt{google/gemma-2-9b-it} significantly outperformed all its RoGemma2-9B fine-tuned variants (including DPO versions) across all few-shot settings, achieving an average accuracy of 47.13\% compared to averages around 39-42\% for the RoGemma2 variants. A similar, though less pronounced, trend was observed for Gemma-1.1-7b.
\end{itemize}
These findings suggest that for complex grammatical reasoning tasks like those in our dataset, the benefits of general multilingual pre-training in strong base models (like Llama 3.1 Instruct or Gemma-2-9B-it) might outweigh the current Romanian-specific fine-tuning approaches applied in the RoLLM project, at least for zero-shot and few-shot settings. The fine-tuning might be more beneficial for other NLP tasks or require different strategies (e.g., incorporating grammatical knowledge more explicitly) to show significant gains on this benchmark. The relatively low overall scores for both base and fine-tuned models (excluding Gemini) further emphasize the difficulty of this task.

\section{Qualitative Analysis: Explanations, Categorization, and Dataset Challenges}
\label{sec:qualitative_analysis}

Beyond quantitative accuracy, understanding LLM capabilities for educational use requires qualitative assessment. This section presents findings from an in-depth analysis focusing primarily on explanations and category classifications generated by \textbf{Gemini 2.5 Pro Experimental} (using CoT prompting). The analysis, conducted by a linguistic expert, evaluated 200 explanations in detail and reviewed LLM category assignments across the full dataset (1151 items). This revealed insights into LLM reasoning, limitations in handling linguistic nuances, and characteristics of the benchmark dataset itself.

The involvement of a linguistic expert was essential to the present study due to the complex and nuanced nature of the materials under analysis. The apparent uniformity of item structure did not preclude the presence of subtle or highly context-dependent linguistic phenomena. Certain errors identified in the dataset were marked by fine semantic or syntactic distinctions that would likely elude detection without expert philological judgment, such as putting the error into the correct language category (see questions 51, 77, 211, 228, etc. where the lexical errors are considered morphological and the opposite) or presenting the explanation in an inadequate style or erroneously.

\subsection{Quality of LLM-Generated Explanations}
While CoT prompting improved answer accuracy (Table~\ref{tab:cot_comparison}), the quality of the accompanying explanations varied significantly. A detailed review of 200 explanations revealed that \textbf{48\% (96 explanations) were problematic} (incomplete, incorrect, misleading, or containing flaws). Key issues included:
\begin{itemize}
    \item \textbf{Incorrect Justifications (67/96):} Explanations frequently provided factually incorrect grammatical reasoning, even when the selected MCQ answer was correct.
    \item \textbf{Imprecision and Irrelevance (29/96):} Explanations often suffered from redundancy, imprecise terminology, irrelevant points, or incoherent concluding statements that undermined the reasoning (e.g., the concluding phrase for Question 26 regarding pronoun agreement).
\end{itemize}
This high error rate in explanations limits the reliability of current LLMs as standalone grammar tutors without expert validation.

\subsection{LLM Performance on Linguistic Categorization}
We also evaluated Gemini 2.5 Pro Experimental's ability to classify each of the 1151 questions into four linguistic domains: Lexical, Morphological, Syntactic, or Phonetic. While the overall accuracy on this task was very high (over 99\% agreement with expert classification), the \textbf{five instances of misclassification} are revealing:
\begin{itemize}
    \item \textit{Morphology vs. Lexicon:} Question 51 (abbreviation of \textit{dumneavoastră}) was misclassified as "Lexical" instead of "Morphological". Question 77 (use of \textit{optim} creating a pleonasm) was misclassified as "Morphological" instead of "Lexical-Semantic".
    \item \textit{Syntax vs. Morphology:} Question 211 (morphological value of pronoun \textit{ce}) and Question 228 (identifying an interjection) were misclassified as "Syntactic". Question 252 (focusing on the preposition \textit{după}) was also misclassified as "Syntactic" instead of "Morphological".
\end{itemize}
These errors typically occurred in borderline cases involving multiple linguistic layers (form, function, meaning), suggesting LLMs may struggle to pinpoint the primary linguistic concept being tested beyond surface cues.

\subsection{Dataset Characteristics and Normative Alignment}
The qualitative analysis also highlighted characteristics and potential issues within the source materials:
\begin{itemize}
    \item \textbf{Unclear/Flawed Prompts:} Seven questions out of the 200 deeply analyzed contained ambiguities or errors affecting comprehensibility and reliable evaluation.
    \item \textbf{Incorrect/Incomplete Official Answers:} Four analyzed items had designated correct answers conflicting with current standards.
    \newpage
    \item \textbf{DOOM 3 Misalignment:} \footnote{DOOM3, published in 2021, serves as the current standard reference for Romanian spelling (ortografic), pronunciation (ortoepic), and morphology (morfologic). While the core orthographic, ortoepic, and morphological norms largely remain unchanged, DOOM3 introduces over 3,000 discrete normative updates. These include the addition or removal of definite/article forms, select flexional variants, and adjustments to accentuation in compound words. It intentionally curtails the proliferation of orthographic variants, helping users to clearly identify the recommended form. It represents the official norms, reason why explanations such as “DOOM 3 prefers” or “recommends” from questions 733 or 896 are not adequate.}This was a notable issue. Analysis of the 19 DOOM-related questions in the dataset revealed:
        \begin{itemize}
            \item \textit{LLM Errors:} In six explanations for DOOM-related questions, Gemini 2.5 Pro provided false or incorrect information regarding DOOM 3 \citep{DOOM3:2021} norms. Examples include incorrectly stating DOOM 3 "accepts" both forms for \textit{cofeina/cafeina} (Q626) or \textit{biscuit/biscuite} (Q1128), misstating the preferred form for \textit{barocă} (Q733), inaccurately describing the abbreviation rule for \textit{dumneavoastră} (Q51), or misrepresenting the recommendation for \textit{cuvânt-înainte} (Q896).
            \item \textit{Outdated Source Questions:} Some source questions themselves appeared outdated relative to DOOM 3 (2021 edition). For example, Question 144 (plural of \textit{proroc}) lacked the standard DOOM 3 form among its options, suggesting the question predates the latest normative update.
        \end{itemize}
\end{itemize}
These findings underscore the need for careful curation and normative alignment of benchmark datasets, especially when evaluating performance on rule-based linguistic tasks.

\subsection{Implications}
The qualitative results emphasize the need for evaluation metrics beyond accuracy, particularly when assessing LLMs for educational roles requiring explanations. The inconsistencies found highlight the importance of dataset quality control and alignment with current linguistic standards (like DOOM 3) for reliable benchmarking in high-stakes domains like language assessment.

\section{Application}

In order to explore the applicability of the created dataset and the generated explanations, we developed a publicly accessible web application 
designed as a learning tool for Romanian language proficiency.

\begin{figure}[H]
  \centering
  \includegraphics[width=\linewidth]{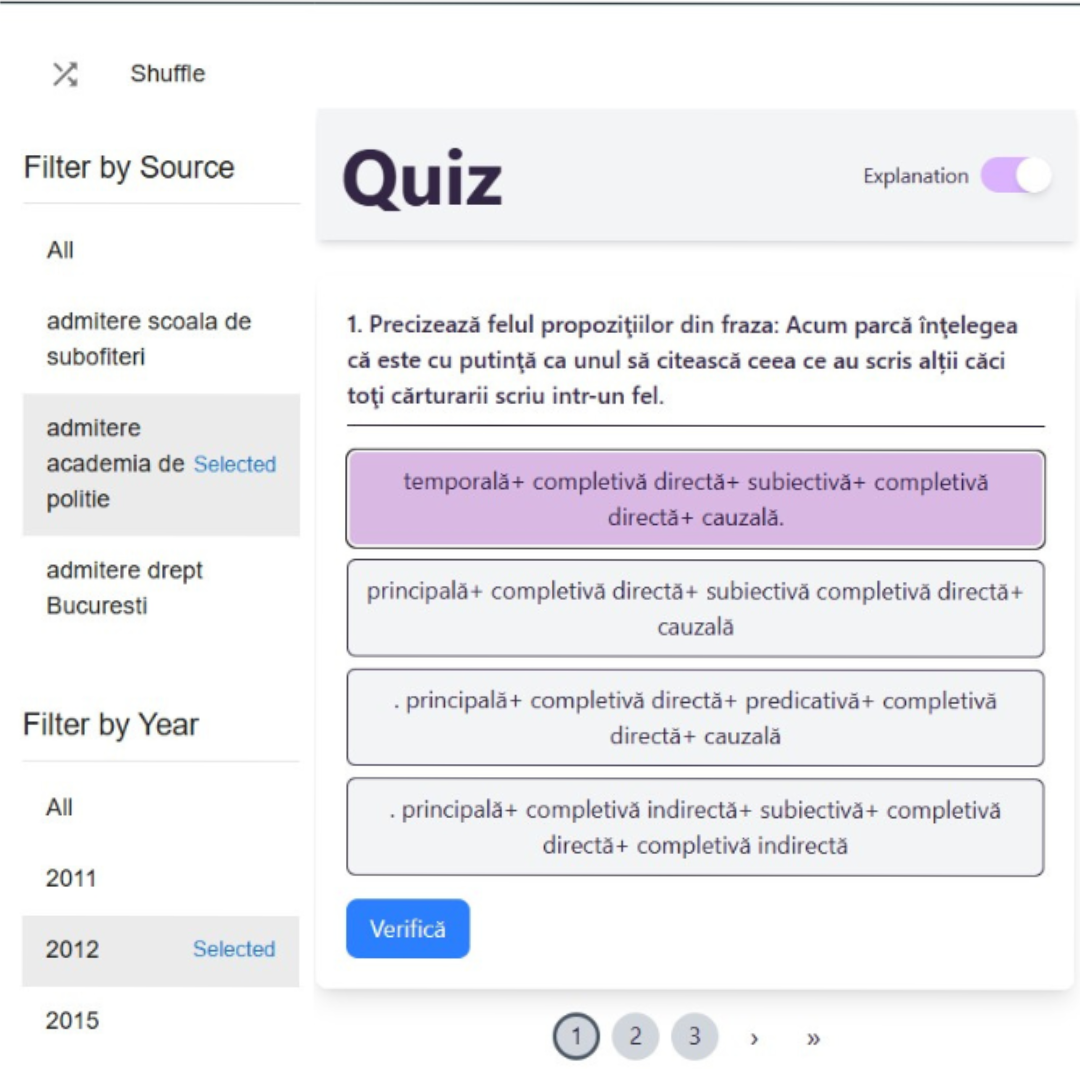} 
  \caption{User interface of the web application, allowing practice with dataset questions.}
  \label{fig:snippet_app_home} 
\end{figure}

The application utilizes the benchmark dataset as its question bank. Users can navigate questions filtered by source (allowing targeted practice based on exam type/difficulty proxy) and submit their answers within a timed or practice mode. Upon submission, the application reveals the correct answer and, importantly, displays the corresponding \textbf{LLM-generated grammatical explanation} (specifically, those generated by Gemini 2.5 Pro and validated for quality by our linguistic expert, as discussed in Section~\ref{sec:qualitative_analysis}).

\begin{figure}[H]
  \centering
  \includegraphics[width=\linewidth]{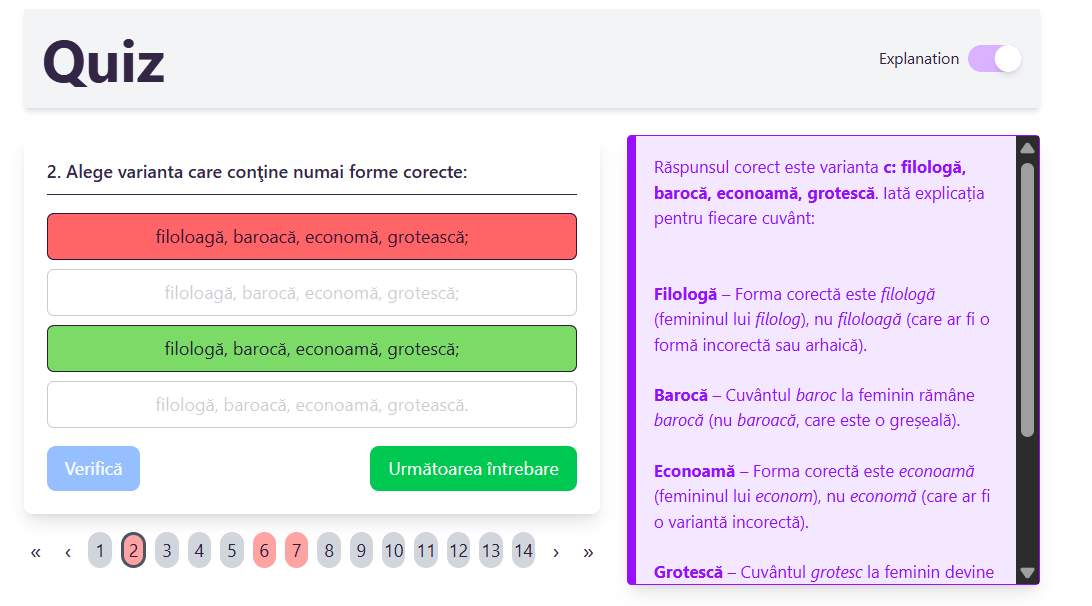} 
  \caption{User interface of the web application, displaying explanations for the correct answers.}
  \label{fig:quiz_explanation} 
\end{figure}

This feature allows learners to not only test their knowledge but also receive immediate feedback and reasoning, leveraging the LLM's explanatory capabilities in a practical educational context. Although explanation quality still requires ongoing validation, providing expert-vetted explanations offers significantly more pedagogical value than simple answer keys. The application is implemented in React and hosted statically, ensuring easy access and deployment.

\section{Conclusion and Future Work}

This study evaluated state-of-the-art LLMs on a novel dataset of Romanian language MCQs sourced from high-stakes exams, focusing on answer accuracy and explanation quality. Our benchmarking, utilizing direct and Chain-of-Thought (CoT) prompting, revealed significant performance variations and a considerable gap between most LLMs and proficient human performance, although \texttt{Gemini 2.5 Pro Experimental} achieved high accuracy (~83\%) using CoT.

Qualitative analysis of Gemini 2.5 Pro's explanations, while revealing flaws in nearly half, also found that the majority were linguistically valuable and useful as a starting point for pedagogical feedback, confirmed by expert assessment. This potential was demonstrated through their integration into our publicly available educational web application, providing learners with practice opportunities augmented by AI-generated, expert-validated grammatical reasoning. However, the analysis also highlighted challenges related to dataset quality and normative alignment (e.g., with DOOM 3).

Our findings underscore the need for both quantitative and qualitative metrics when evaluating LLMs for educational roles, particularly concerning explanation generation in low-resource languages. While current LLMs show promise as assistive tools, significant work remains to ensure consistent accuracy and pedagogical soundness.

\textbf{Future work} should proceed along several key directions. Firstly, \textbf{expanding the benchmark scope} is essential: incorporating more questions, annotations (like difficulty levels or specific grammar subsections), evaluating a broader range of SOTA models (including newer Romanian-specific ones), and performing detailed cross-lingual analyses. Secondly, significant potential lies in \textbf{enhancing model performance and explanation quality} through advanced techniques. This includes exploring \textbf{Retrieval-Augmented Generation (RAG)} leveraging Romanian language manuals or grammar resources as external knowledge, targeted \textbf{fine-tuning} using a subset of the dataset questions, and experimenting with more sophisticated prompting strategies such as \textbf{few-shot learning} to potentially create more effective AI tutors. Finally, addressing the identified dataset limitations through continued \textbf{curation and normative alignment} remains vital for building truly robust evaluation benchmarks for Romanian educational NLP.

\section{Limitations}
\label{sec:limitations}

Our dataset of 1\,151 Romanian multiple-choice questions (MCQs) is smaller than the large benchmarks available for high-resource languages and represents mainly high-stakes exam material. Items were extracted with Tesseract OCR and then spot-checked, yet some annotation noise may remain. Question categorisation relied on an LLM-assisted workflow that, despite manual review, can still inherit model errors.

While we have categorized the questions when answering them and obtained the following split syntactic 382 questions, morphological 364, lexical 319 and phonetic 86, we have not annotated the initial data with this information.

Methodologically, the study is restricted to MCQs; open-ended generation and deeper discourse skills are outside our present scope. 

\section{Ethical Considerations}
\label{sec:ethics}

This research uses publicly available Romanian multiple-choice exam questions; all sources are free of licensing or privacy constraints.  
We release the dataset on Hugging Face to foster transparency, reproducibility, and further work on low-resource educational NLP.  
The free web application stores no PII.  
Dataset content mirrors the focus of the original exams.  
LLM-generated explanations were validated by a linguist to prevent misinformation.  
We acknowledge possible training-data biases and welcome community scrutiny; users should respect original copyrights.


\bibliography{custom}

\begin{thebibliography}{20}
\expandafter\ifx\csname natexlab\endcsname\relax\def\natexlab#1{#1}\fi

\bibitem[{Conneau et~al.(2020)Conneau, Khandelwal, Goyal, Chaudhary, Wenzek, Guzm{\'a}n, Grave, Ott, Zettlemoyer, and Stoyanov}]{conneau-etal-2020-unsupervised}
Alexis Conneau, Kartikay Khandelwal, Naman Goyal, Vishrav Chaudhary, Guillaume Wenzek, Francisco Guzm{\'a}n, Edouard Grave, Myle Ott, Luke Zettlemoyer, and Veselin Stoyanov. 2020.
\newblock \href {https://doi.org/10.18653/v1/2020.acl-main.747} {Unsupervised cross-lingual representation learning at scale}.
\newblock In \emph{Proceedings of the 58th Annual Meeting of the Association for Computational Linguistics}, pages 8440--8451, Online. Association for Computational Linguistics.

\bibitem[{Cui et~al.(2024)Cui, Yang, and Yao}]{cui2023chinese}
Yiming Cui, Ziqing Yang, and Xin Yao. 2024.
\newblock \href {http://arxiv.org/abs/2304.08177} {Efficient and effective text encoding for chinese llama and alpaca}.

\bibitem[{DeepSeek-AI et~al.(2025)DeepSeek-AI, Liu, Feng, Xue, Wang, Wu, Lu, Zhao, Deng et~al.}]{liu2025deepseek}
DeepSeek-AI, Aixin Liu, Bei Feng, Bing Xue, Bingxuan Wang, Bochao Wu, Chengda Lu, Chenggang Zhao, Chengqi Deng, et~al. 2025.
\newblock \href {http://arxiv.org/abs/2412.19437} {Deepseek-v3 technical report}.

\bibitem[{Grattafiori et~al.(2024)Grattafiori, Dubey, Jauhri, Pandey, Kadian, Al-Dahle, Letman, Mathur, Schelten, Vaughan et~al.}]{grattafiori2024llama}
Aaron Grattafiori, Abhimanyu Dubey, Abhinav Jauhri, Abhinav Pandey, Abhishek Kadian, Ahmad Al-Dahle, Aiesha Letman, Akhil Mathur, Alan Schelten, Alex Vaughan, et~al. 2024.
\newblock \href {http://arxiv.org/abs/2407.21783} {The llama 3 herd of models}.

\bibitem[{Hendrycks et~al.(2021)Hendrycks, Burns, Basart, Zou, Mazeika, Song, and Steinhardt}]{hendrycks2021measuringmassivemultitasklanguage}
Dan Hendrycks, Collin Burns, Steven Basart, Andy Zou, Mantas Mazeika, Dawn Song, and Jacob Steinhardt. 2021.
\newblock \href {http://arxiv.org/abs/2009.03300} {Measuring massive multitask language understanding}.

\bibitem[{Hu et~al.(2020)Hu, Ruder, Siddhant, Neubig, Firat, and Johnson}]{hu2020xtreme}
Junjie Hu, Sebastian Ruder, Aditya Siddhant, Graham Neubig, Orhan Firat, and Melvin Johnson. 2020.
\newblock \href {https://arxiv.org/abs/2003.11080} {Xtreme: A massively multilingual multi-task benchmark for evaluating cross-lingual generalization}.
\newblock In \emph{Proceedings of the 37th International Conference on Machine Learning}.

\bibitem[{{Institutul de Lingvistic\u{a} ``Iorgu Iordan -- Alexandru Rosetti''}(2021)}]{DOOM3:2021}
{Institutul de Lingvistic\u{a} ``Iorgu Iordan -- Alexandru Rosetti''}. 2021.
\newblock \emph{Dic\c{t}ionarul Ortografic, Ortoepic \c{s}i Morfologic al Limbii Rom\^{a}ne ({DOOM})}, 3 edition.
\newblock Editura Univers Enciclopedic, Bucure\c{s}ti.
\newblock Coordonatoare: Ioana Vintil\u{a}-R\u{a}dulescu.

\bibitem[{Kuratov and Arkhipov(2019)}]{kuratov-arkhipov-2019-adapting}
Yuri Kuratov and Mikhail Arkhipov. 2019.
\newblock \href {http://arxiv.org/abs/1905.07213} {Adaptation of deep bidirectional multilingual transformers for russian language}.

\bibitem[{L{\'o}pez~Cortez et~al.(2024)L{\'o}pez~Cortez, Norris, and Duman}]{lopez-cortez-etal-2024-gmeg}
S.~Magal{\'i} L{\'o}pez~Cortez, Mark~Josef Norris, and Steve Duman. 2024.
\newblock \href {https://aclanthology.org/2024.lrec-main.688/} {{GMEG}-{EXP}: A dataset of human- and {LLM}-generated explanations of grammatical and fluency edits}.
\newblock In \emph{Proceedings of the 2024 Joint International Conference on Computational Linguistics, Language Resources and Evaluation (LREC-COLING 2024)}, pages 7785--7800, Torino, Italia. ELRA and ICCL.

\bibitem[{Luukkonen et~al.(2023)Luukkonen, Komulainen, Luoma, Eskelinen, Kanerva, Kupari, Ginter, Laippala, Muennighoff, Piktus, Wang, Tazi, Scao, Wolf, Suominen, Sairanen, Merioksa, Heinonen, Vahtola, Antao, and Pyysalo}]{luukkonen-etal-2023-fingpt}
Risto Luukkonen, Ville Komulainen, Jouni Luoma, Anni Eskelinen, Jenna Kanerva, Hanna-Mari Kupari, Filip Ginter, Veronika Laippala, Niklas Muennighoff, Aleksandra Piktus, Thomas Wang, Nouamane Tazi, Teven Scao, Thomas Wolf, Osma Suominen, Samuli Sairanen, Mikko Merioksa, Jyrki Heinonen, Aija Vahtola, Samuel Antao, and Sampo Pyysalo. 2023.
\newblock \href {https://doi.org/10.18653/v1/2023.emnlp-main.164} {{F}in{GPT}: Large generative models for a small language}.
\newblock In \emph{Proceedings of the 2023 Conference on Empirical Methods in Natural Language Processing}, pages 2710--2726, Singapore. Association for Computational Linguistics.

\bibitem[{Martin et~al.(2020)Martin, Muller, Ortiz~Su{\'a}rez, Dupont, Romary, de~la Clergerie, Seddah, and Sagot}]{martin-etal-2020-camembert}
Louis Martin, Benjamin Muller, Pedro~Javier Ortiz~Su{\'a}rez, Yoann Dupont, Laurent Romary, {\'E}ric de~la Clergerie, Djam{\'e} Seddah, and Beno{\^i}t Sagot. 2020.
\newblock \href {https://doi.org/10.18653/v1/2020.acl-main.645} {{C}amem{BERT}: a tasty {F}rench language model}.
\newblock In \emph{Proceedings of the 58th Annual Meeting of the Association for Computational Linguistics}, pages 7203--7219, Online. Association for Computational Linguistics.

\bibitem[{Masala et~al.(2024)Masala, Ilie-Ablachim, Corlatescu, Zavelca, Leordeanu, Velicu, Popescu, Dascalu, and Rebedea}]{masala2024openllmrotechnicalreport}
Mihai Masala, Denis~C. Ilie-Ablachim, Dragos Corlatescu, Miruna Zavelca, Marius Leordeanu, Horia Velicu, Marius Popescu, Mihai Dascalu, and Traian Rebedea. 2024.
\newblock \href {http://arxiv.org/abs/2405.07703} {Openllm-ro -- technical report on open-source romanian llms}.

\bibitem[{Masala et~al.(2020)Masala, Ruseti, and Dascalu}]{masala-etal-2020-robert}
Mihai Masala, Stefan Ruseti, and Mihai Dascalu. 2020.
\newblock \href {https://doi.org/10.18653/v1/2020.coling-main.581} {{R}o{BERT} {--} a {R}omanian {BERT} model}.
\newblock In \emph{Proceedings of the 28th International Conference on Computational Linguistics}, pages 6626--6637, Barcelona, Spain (Online). International Committee on Computational Linguistics.

\bibitem[{Mayor-Rocher et~al.(2024{\natexlab{a}})Mayor-Rocher, Melero, Merino-Gómez, González, Ferrando, Conde, and Reviriego}]{mayor-rocher2024teleiaeval}
M.~Mayor-Rocher, N.~Melero, E.~Merino-Gómez, M.~González, R.~Ferrando, J.~Conde, and P.~Reviriego. 2024{\natexlab{a}}.
\newblock \href {https://doi.org/10.48550/arXiv.2409.15334} {Evaluating large language models with tests of spanish as a foreign language: Pass or fail?}

\bibitem[{Mayor-Rocher et~al.(2024{\natexlab{b}})Mayor-Rocher, Melero, Merino-Gómez, González, Ferrando, Conde, and Reviriego}]{spanish_benchmark_teleia}
M.~Mayor-Rocher, N.~Melero, E.~Merino-Gómez, M.~González, R.~Ferrando, J.~Conde, and P.~Reviriego. 2024{\natexlab{b}}.
\newblock \href {https://doi.org/10.5281/zenodo.12571763} {{Spanish Language Benchmark for Artificial Intelligence Models (TELEIA)}}.

\bibitem[{Nagata(2019)}]{nagata2019toward}
Ryo Nagata. 2019.
\newblock \href {https://doi.org/10.18653/v1/D19-1316} {Toward a task of feedback comment generation for writing learning}.
\newblock In \emph{Proceedings of the 2019 Conference on Empirical Methods in Natural Language Processing and the 9th International Joint Conference on Natural Language Processing (EMNLP-IJCNLP)}, pages 3206--3215, Hong Kong, China. Association for Computational Linguistics.

\bibitem[{OpenAI(2024)}]{openai2024gpt4technicalreport}
OpenAI. 2024.
\newblock \href {http://arxiv.org/abs/2303.08774} {Gpt-4 technical report}.

\bibitem[{P{\u{a}}i{\c{s}} and Tufi{\c{s}}(2023)}]{Pais2023}
Vasile P{\u{a}}i{\c{s}} and Dan Tufi{\c{s}}. 2023.
\newblock \href {https://doi.org/10.1007/978-3-031-28819-7_31} {\emph{Language Report Romanian}}, pages 199--202. Springer International Publishing, Cham.

\bibitem[{Team et~al.(2024)Team, Anil, Borgeaud, Alayrac, Yu, Soricut, Schalkwyk, Dai, Hauth, Millican, Silver et~al.}]{geminiteam2024geminifamilyhighlycapable}
Gemini Team, Rohan Anil, Sebastian Borgeaud, Jean-Baptiste Alayrac, Jiahui Yu, Radu Soricut, Johan Schalkwyk, Andrew~M. Dai, Anja Hauth, Katie Millican, David Silver, et~al. 2024.
\newblock \href {http://arxiv.org/abs/2312.11805} {Gemini: A family of highly capable multimodal models}.

\bibitem[{Yang et~al.(2025)Yang, Yang, Zhang, Hui, Zheng, Yu, Li, Liu, Huang, Wei et~al.}]{yang2024qwen2}
An~Yang, Baosong Yang, Beichen Zhang, Binyuan Hui, Bo~Zheng, Bowen Yu, Chengyuan Li, Dayiheng Liu, Fei Huang, Haoran Wei, et~al. 2025.
\newblock \href {http://arxiv.org/abs/2412.15115} {Qwen2. 5 technical report}.
\newblock \emph{arXiv preprint arXiv:2412.15115}.

\end{thebibliography}

\end{document}